\documentclass[lettersize,journal,dvipsnames]{IEEEtran}
\usepackage{amsmath,amsfonts}
\usepackage{algorithmic}
\usepackage{algorithm}
\usepackage{array}
\usepackage[caption=false,font=normalsize,labelfont=sf,textfont=sf]{subfig}
\usepackage{textcomp}
\usepackage{stfloats}
\usepackage{url}
\usepackage{verbatim}
\usepackage{graphicx}
\usepackage{cite}
\usepackage{enumitem}
\usepackage{svg}
\usepackage{balance}
\usepackage{scalerel}
\usepackage{ulem}

\usepackage[para]{threeparttable}
\usepackage{booktabs}
\usepackage{multirow}
\usepackage{multicol}

\usepackage{xcolor}
\usepackage[theorems,skins]{tcolorbox}
\usepackage{empheq}
\tcbset{red eqbox/.style={enhanced,top=0.05ex, bottom=0.05ex, 
left=0.1ex,right=0.1ex,
overlay={\fill[red!10] (frame.south west) to[bend left] 
 (frame.north west) --  (frame.north east) to[bend left]
  (frame.south east) -- cycle;},
boxrule=0pt},
blue eqbox/.style={enhanced,top=0ex, bottom=0ex, 
left=0ex,right=0ex,
overlay={\fill[blue!10] (frame.south west) to[bend left] 
 (frame.north west) --  (frame.north east) to[bend left]
  (frame.south east) -- cycle;},
boxrule=0pt},
blue1 eqbox/.style={enhanced,top=0ex, bottom=0ex, 
left=0ex,right=0ex,
overlay={\fill[blue!5] (frame.south west) to[bend left] 
 (frame.north west) --  (frame.north east) to[bend left]
  (frame.south east) -- cycle;},
boxrule=0pt},
blue2 eqbox/.style={enhanced,top=0ex, bottom=0ex, 
left=0ex,right=0ex,
overlay={\fill[blue!10] (frame.south west) to[bend left] 
 (frame.north west) --  (frame.north east) to[bend left]
  (frame.south east) -- cycle;},
boxrule=0pt},
blue3 eqbox/.style={enhanced,top=0ex, bottom=0ex, 
left=0ex,right=0ex,
overlay={\fill[blue!15] (frame.south west) to[bend left] 
 (frame.north west) --  (frame.north east) to[bend left]
  (frame.south east) -- cycle;},
boxrule=0pt},
blue4 eqbox/.style={enhanced,top=0ex, bottom=0ex, 
left=0ex,right=0ex,
overlay={\fill[blue!20] (frame.south west) to[bend left] 
 (frame.north west) --  (frame.north east) to[bend left]
  (frame.south east) -- cycle;},
boxrule=0pt},
blue5 eqbox/.style={enhanced,top=0ex, bottom=0ex, 
left=0ex,right=0ex,
overlay={\fill[blue!25] (frame.south west) to[bend left] 
 (frame.north west) --  (frame.north east) to[bend left]
  (frame.south east) -- cycle;},
boxrule=0pt},
blue6 eqbox/.style={enhanced,top=0ex, bottom=0ex, 
left=0ex,right=0ex,
overlay={\fill[blue!30] (frame.south west) to[bend left] 
 (frame.north west) --  (frame.north east) to[bend left]
  (frame.south east) -- cycle;},
boxrule=0pt},
highlight math style=red eqbox,
}

\begin{document}

\title{SenDaL: An Effective and Efficient Calibration Framework of Low-Cost Sensors for Daily Life}

% \def\dsp{\def\baselinestretch{1.02}\large\normalsize}
% \dsp
% \def\ffsp{\def\baselinestretch{1.02}\large\normalsize}
% \ffsp

\author{Seokho Ahn, Hyungjin Kim, Euijong Lee, Young-Duk Seo
% <-this % stops a space
\thanks{Manuscript received 13 September 2023; revised 3 February 2024; accepted 25 February 2024. Date of publication 28 February 2024; date of current
version 23 May 2024. 
This work was supported in part by the National Research Foundation of Korea (NRF) Grant funded by the Korea Government (MSIT) under Grant NRF-2022R1C1C1012408, and in part by the Institute of Information \& Communications Technology Planning \& Evaluation (IITP) Grant funded by the Korea Government (MSIT) 
through Artificial Intelligence Convergence Innovation Human Resources Development (Inha University) under Grant RS-2022-00155915 and through Deep Total Recall: Continual Learning for Human-Like Recall of Artificial Neural Networks (10\%) under Grant 2022-0-00448. (\textit{Corresponding author}: \textit{Young-Duk Seo}.)}
\thanks{Seokho Ahn, Hyungjin Kim, and Young-Duk Seo are with the Department of Electrical and Computer Engineering, Inha University, Incheon 22212, South Korea (e-mail: sokho0514@gmail.com; flslzk@gmail.com; mysid88@inha.ac.kr).}
\thanks{Euijong Lee is with the School of Computer Science, Chungbuk National University, Cheongju-si 28644, South Korea (e-mail: kongjjagae@cbnu.ac.kr).}
\thanks{Digital Object Identifier 10.1109/JIOT.2024.3371150}
}

\markboth{IEEE Internet of Things Journal,~Vol.~11, No.~11, 1~JUNE~2024}%
{Ahn \MakeLowercase{\textit{et al.}}: SenDaL: An Effective and Efficient Calibration Framework of Low-cost Sensors for Daily Life}


\maketitle
\begin{abstract}
The collection of accurate and noise-free data is a crucial part of Internet of Things (IoT)-controlled environments. However, the data collected from various sensors in daily life often suffer from inaccuracies. Additionally, IoT-controlled devices with low-cost sensors lack sufficient hardware resources to employ conventional deep learning models. To overcome this limitation, we propose SenDaL (\textbf{Sen}sors for \textbf{Da}ily \textbf{L}ife), the first framework that utilizes neural networks for calibrating low-cost sensors. SenDaL introduces novel training and inference processes that enable it to achieve accuracy comparable to deep learning models while simultaneously preserving latency and energy consumption similar to linear models. SenDaL is first trained in a bottom-up manner, making decisions based on calibration results from both linear and deep learning models. Once both models are trained, SenDaL makes independent decisions through a top-down inference process, ensuring accuracy and inference speed. Furthermore, SenDaL can select the optimal deep learning model according to the resources of the IoT devices because it is compatible with various deep learning models such as LSTM-based and Transformer-based models. We have verified that SenDaL outperforms existing deep learning models in terms of accuracy, latency, and energy efficiency through experiments conducted in different IoT environments and real-life scenarios.
\end{abstract}

\begin{IEEEkeywords}
Bottom-up training, deep learning, Internet of Things (IoT), sensor calibration, soft sensor, top-down inference.
\end{IEEEkeywords}

\section{Introduction}
\IEEEPARstart{W}{e live} in a daily life where things are interconnected through the Internet of Things (IoT) \cite{Patel_2016, Lee_2021}. Collecting real-time data from various sensors in an IoT environment has significantly improved the quality of our daily lives by providing access to information that would otherwise be challenging to collect. However, the collected data often suffer from low accuracy due to hardware limitations in the sensors themselves. To enhance sensor accuracy, several studies in the fields of data-driven robotics \cite{Gupta_2018, Serkan_2019, Kleeberger_2020, Barclay_2022} and soft-sensors \cite{Ke_2017, LoyBenitez_2020, Curreri_2021, Sun_2021, Andrade_2022} have calibrated the sensor data using deep learning techniques and demonstrated remarkable performance. Despite these advancements, there remains a research gap concerning the enhancement of low-cost sensors used in daily life, such as fine-dust, temperature, and humidity sensors, by applying deep learning.

Improving the performance of low-cost sensors can be highly beneficial in various real-life scenarios as it enables accurate access to sensor data, contributing to a more informed decision-making process. However, calibrating low-cost sensors used in daily life with high performance can be difficult due to three main challenges:

%====================================================================
\begin{figure}[t]
\centering
\includegraphics[width=0.85\columnwidth]{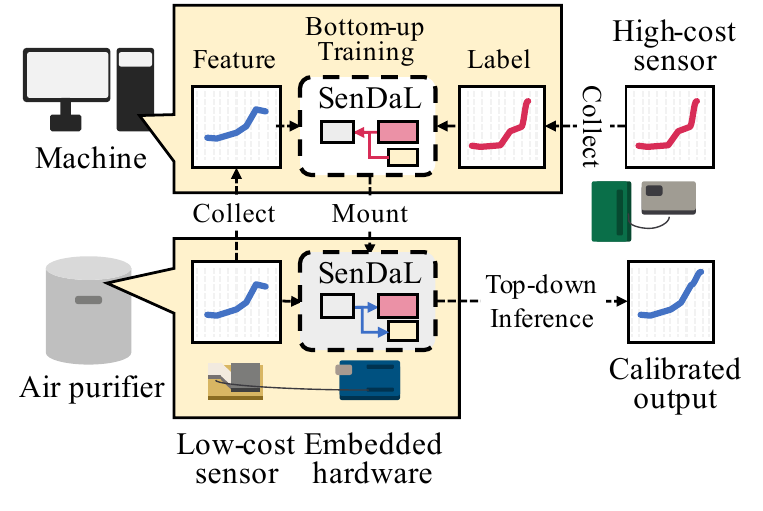}
\caption{\textbf{Overview of SenDaL.} Both high-cost and low-cost sensors are used to train SenDaL on a machine through a bottom-up training process. Then, a top-down inference process enables SenDaL calibrate low-cost sensors in IoT-controlled devices with high accuracy and fast inference speed. }
\label{intro}
\end{figure}
%====================================================================

\begin{itemize}[noitemsep, leftmargin=1.05em]
\item \textbf{\textit{Model accuracy}}: The model must be capable of enhancing the accuracy of low-cost sensors, ensuring reliability and trustworthiness. Moreover, this improvement in accuracy needs to be achieved with a low computational cost.
\item \textbf{\textit{Constrained resources}}: IoT devices used in daily-life systems have limited available hardware resources, such as CPU and memory. Hence, the task should operate effectively within these constraints to ensure optimal performance.
\item \textbf{\textit{Real-time inference}}: An effective monitoring tool should be able to provide post-processed data in real-time. This necessitates efficient and fast processing of the collected data.
\end{itemize}

To improve low-cost sensors used in daily life, we propose SenDaL (\textbf{Sen}sors for \textbf{Da}ily \textbf{L}ife), the first calibration framework based on neural networks for low-cost sensors that present accurate data in real-time within limited resources. An overview of SenDaL is presented in Figure 1. 

SenDaL operates on two different processes: the \textit{bottom-up training} (detailed in Section \ref{sec_training}) and \textit{top-down inference} (detailed in Section \ref{sec_inference}) processes in order to enable effective training and efficient inference. The bottom-up training process converges the calibration results of two calibration models, such as linear and deep learning models, to produce more accurate calibration results than existing deep learning models \cite{Hochreiter_1997, Cho_2014, Neil_2016, Vaswani_2017, Kitaev_2020, Choromanski_2021}. After having the ability to select two calibration models through a bottom-up training process, the top-down inference process selectively uses these two models. Therefore, SenDaL achieves accuracy comparable to deep learning models while maintaining energy consumption and inference speed close to that of the linear model.

In summary, SenDaL demonstrates faster inference and improved accuracy in real-life scenarios compared to existing studies \cite{Domb_19, Cheng_2020} that solely focus on enhancing accuracy. To verify the performance of SenDaL, we utilize fine-dust sensors that are frequently used in daily life, and fine-dust data requires accurate calibration models due to its significant variations.

Our contributions in this study are as follows: 
\begin{itemize}[noitemsep, leftmargin=1.05em]
\item We propose SenDaL, the first framework for calibrating low-cost sensors in IoT environments, which aims to improve the accuracy to be close to high-cost sensors.

\item SenDaL employs bottom-up training and top-down inference processes to improve efficiency and make it much faster than the baseline deep learning models.

\item We design SenDaL to be flexible in various IoT environments, allowing for the selection of various deep learning models to fit given hardware resources. 

\item We conduct experiments with our refined dataset, and SenDaL outperforms the baseline deep learning models on various embedded hardware in most cases.

\end{itemize}

The remainder of this paper is organized as follows. First, we represent related works in Section \ref{sec_related} and define the problem statement in Section \ref{sec_problem}. Next, we explain the SenDaL framework focusing on the bottom-up training and top-down inference processes in detail in Section \ref{sec_model}. In Section \ref{sec_experiment}, we conduct the experiment and verify the superiority of our framework. Section \ref{sec_discussion} addresses the limitations of our studies and outlines plans for potential solutions. The final section discusses the conclusions and future studies.

\section{Related Work\label{sec_related}}
\subsection{Deep Learning-based Time-series Prediction}
Advancements in deep learning have led to the development of time-series forecasting models, such as LSTM \cite{Hochreiter_1997} and Transformer \cite{Vaswani_2017}, that provide more accurate predictions than other techniques. The method of calibrating time-series data is similar to that of forecasting time-series. Therefore, it is essential to evaluate existing deep learning-based forecasting models to identify an appropriate technique for calibrating data acquired from low-cost sensors.

Several studies have aimed to minimize the necessary hardware resources for prediction. In particular, variations of the LSTM model, such as GRU \cite{Cho_2014} and C-LSTM \cite{Wang_2018}, have been proposed to improve the speed of predictions by modifying the structure of the traditional LSTM. Other methods, such as those proposed by \cite{Hoyer_2022} and \cite{Chen_2022}, have been developed to enhance efficiency without changing the LSTM structure. For Transformer-based models, new architectures such as Informer \cite{Zhou_2020}, Reformer \cite{Kitaev_2020}, Performer \cite{Choromanski_2021}, Pyraformer \cite{Liu_2021}, and Ecoformer \cite{Liu_2022} have been proposed to improve the time and space complexity of the attention mechanism. These studies enable the calibration of low-cost sensors using deep learning models on reduced hardware resources. However, compared to the linear model, the resource utilization is considerably higher, but the improvement in calibration accuracy is not significant.

Some studies, such as Phased LSTM \cite{Neil_2016} and THP \cite{Zuo_2020}, have focused on event time-series. Forecasting events is crucial since the data collected from sensors exhibits non-periodic characteristics. However, they are not suitable for low-cost sensor calibration due to high computational complexity.

\subsection{Hybrid Machine Learning}
In the field of machine learning, the term \textit{hybrid machine learning} is used in two distinct contexts. First, it refers to a methodology used to construct a model by simultaneously combining two or more techniques or models, also known as ensemble learning \cite{Oliveira_2014}. This technique exhibits powerful performance for task prediction \cite{Stefenon_2022}, and various ensemble methods are employed for predictions \cite{Gastinger_2021, Zhang_2022, Cawood_2021}. These hybrid models are mainly based on machine learning or statistical techniques \cite{Gastinger_2021, Zhang_2022}, but ensembles using deep learning models are also conducted \cite{Cawood_2021}. Their hybrid methods demonstrate higher accuracy compared to using a single model, however, the use of multiple models leads to slower inference speeds and more hardware utilization, which poses challenges for use in IoT-controlled devices.

Second, hybrid machine learning is a technology that enables the creation of models applicable in multiple environments \cite{Liu_2020}. Most studies focus on reducing model complexity \cite{Liu_2020, Asutkar_2023} or excluding certain operations to support inference on various hardware platforms \cite{David_2020, Aydonat_2017}. However, this lightweight approach does not improve inference speed to the same extent as the linear model, as both training and inference methods remain unchanged. In contrast, SenDaL operates two separate methods, namely bottom-up training and top-down inference, resulting in a significant improvement in latency and energy consumption during the inference process.

\subsection{Data-driven Robotics and Soft-Sensors}
The use of data-driven approaches in robotics has been gradually increasing, since they have been shown to improve performance \cite{Barclay_2022}, such as enhancing the capability of robots to acquire knowledge \cite{Serkan_2019} or improve the accuracy of tasks, such as robot grasping \cite{Gupta_2018, Kleeberger_2020}. However, these methods cannot be applied to low-cost sensors, as they are primarily designed for use in high-cost sensors or robot components.

In soft-sensors, data-driven approaches are used in a wide range of applications, from industry-scale environments to various tasks suitable for IoT environments, using diverse deep learning models \cite{Ke_2017, LoyBenitez_2020, Sun_2021}. Various deep learning models have been used for soft-sensor \cite{Sun_2021}, but LSTM-based models are mainly used in their domains \cite{Ke_2017, LoyBenitez_2020}. Moreover, research has focused on monitoring the sensor reliability or calibrating data based on multiple low-cost sensors \cite{Cheng_2020, Cheng_2022, Zuniga_2022}. Nevertheless, these methods are not sufficiently simple for increasing the accuracy of a single low-cost sensor and do not improve the latency because they do not consider real-time inference. In summary, there is currently no research that comprehensively considers all of the accuracy, execution speed, and energy consumption of a model using only a single low-cost sensor.

\section{Problem Statement \label{sec_problem}}
\subsection{Problem Definition}
We consider two sensors, \(\mathcal{X}\) and \(\mathcal{Y}\), that have the same function in IoT devices. Sensor \(\mathcal{X}\) is more cost-efficient, but less accurate than sensor \(\mathcal{Y}\). Our goal is to improve the accuracy of sensor \(\mathcal{X}\) to that of \(\mathcal{Y}\); formally, we propose a deep learning model \(\mathcal{F}(*)\) for calibrating the \(i^{th}\) observed value \(x_i\) from \(\mathcal{X}\) to match \(y_i\) from \(\mathcal{Y}\) that is regarded as ground truth. The model uses the previous observations of sensor \(\mathcal{X}\), we use the \(i^{th}\) time-series as inputs in our model, defined as follows:
\begin{equation}\label{Timeseries}
    \mathcal{S}_i:=\left(x_j\right)_{\substack{i-N< j\leq i \\ j>0}}
\end{equation}
where \(N\) is the window size used for training the model \(\mathcal{F}\), while the index \(i\) is often omitted when the explanation of \(\mathcal{S}\) is independent of the \(i^{th}\) observation. We aim to minimize the difference between \(\mathcal{F}(\mathcal{S})=\hat{x}\) and the ground truth \(y\), so that the measured data from the updated sensor \(\mathcal{X}\) will be more accurate than the original sensor.

%===================================================================
\begin{figure}[t]
\centering
\includegraphics[width=0.9\columnwidth]{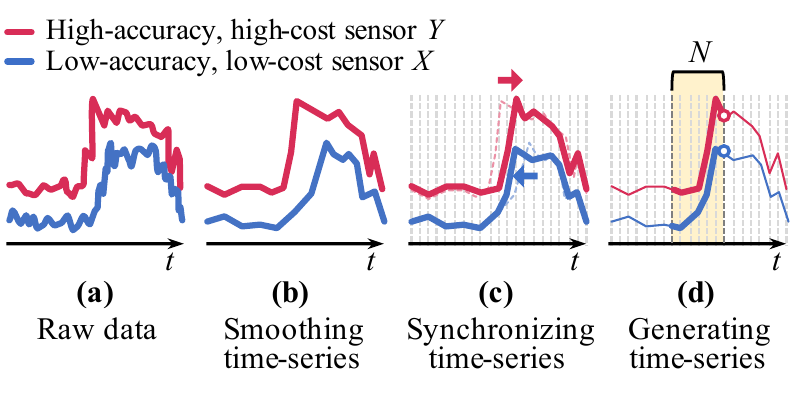} 
\caption{Overall data refinement process.}
\label{refinement}
\end{figure}
%===================================================================

\subsection{Model Requirements}
The model \(\mathcal{F}\) is designed for embedded hardware such as home appliances, thus it must strike a balance between effective model performance and efficient resource utilization. Therefore, the requirements that our model should consider are as follows:

\begin{itemize}[noitemsep, leftmargin=1.2em]
\item \textbf{\textit{Model accuracy}}: The accurate measurement of sensor data is crucial in deriving meaningful decisions in an IoT environment. However, increasing model accuracy while considering real-time inference and hardware resources poses a challenge. While some studies attempt to overcome hardware limitations and improve sensor accuracy \cite{Domb_19, Cheng_2020}, they primarily focus on using linear models, which are prone to errors and result in lower accuracy.

\item \textbf{\textit{Constrained resources}}: Hardware performance encompasses not only the data collection ability of the sensor itself but also the computing resources available for sensor calibration. The calibration of low-cost sensors is much harder than that of high-cost sensors as the data collected by the low-cost sensors includes many outliers \cite{Culic_2020}. Thus, only limited domains (i.e., data-driven robotics \cite{Gupta_2018, Serkan_2019, Kleeberger_2020, Barclay_2022} and soft sensor \cite{Ke_2017, LoyBenitez_2020, Curreri_2021, Sun_2021, Andrade_2022}) that can be studied primarily in large-scale environments such as process industries and smart buildings can calibrate high-cost sensors with sufficient hardware resources.  However, these approaches cannot be directly applicable to upgrading low-cost sensors.

\item \textbf{\textit{Real-time inference}}: To utilize the measured data from IoT home appliances promptly, real-time inference of low-cost sensors is essential. Although some studies focus on improving sensor performance in smaller environments \cite{Ke_2017, Andrade_2022, Ali_2023, JP_2023}, they do not adequately address the need for real-time inference. Achieving real-time calibration results using high-performance models requires additional hardware resources. However, low-cost sensors do not possess sufficient resources for this purpose.

\end{itemize}

In summary, calibrating low-cost sensors used in daily life with high performance can be challenging due to achieving high accuracy within limited hardware resources. Moreover, addressing these challenges with existing methods, such as simply reducing or optimizing the deep learning models \cite{Hoyer_2022, Chen_2022, Zhou_2020, Kitaev_2020, Choromanski_2021, Liu_2021, Liu_2022}, is difficult because calibration model for low-cost sensors requires achieving inference speed and energy consumption levels similar to linear models. Thus, it is essential to develop new training and inference methods that enable model \(\mathcal{F}\) to achieve accuracy comparable to deep learning models, inference speed close to the linear model, and energy consumption similar to the linear model.

%===================================================================
\begin{figure}[t]
\centering
\includegraphics[width=0.7\columnwidth]{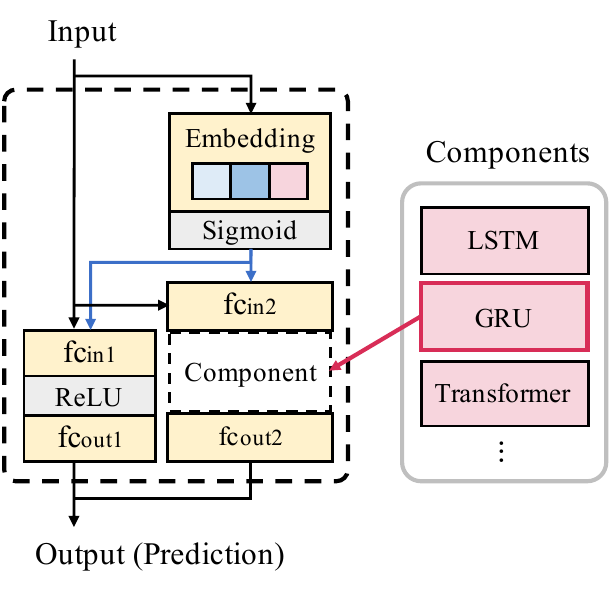} 
\caption{\textbf{Overall structure of SenDaL framework.} SenDaL can perform direct inference through either the linear layer (marked in a blue box) or the component layer (marked in a red box).}
\setlength{\textfloatsep}{3pt}
\label{model}
\end{figure}
%===================================================================

%===================================================================
\begin{figure*}[t]
\centering
\includegraphics[width=0.95\textwidth]{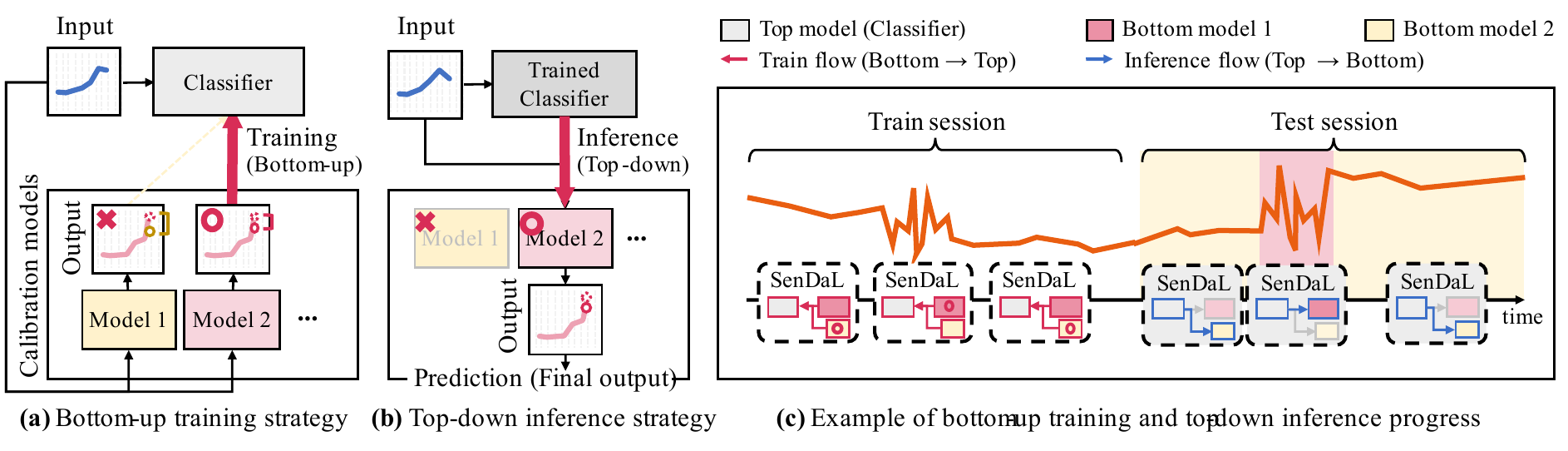} 
\caption{\textbf{Two types of decision-making concept.} \textbf{(a)} Embedding layer (Classifier) learns a model determination method using the results of several calibration models, in a bottom-up manner. \textbf{(b)} Once the embedding layer is trained, the embedding layer employs a top-down strategy during the inference process, independently formulating decisions without assistance from any calibration model. \textbf{(c)} Through different training and inference processes, fast and accurate inference results can be made in an actual hardware environment.  }
\label{concept}
\end{figure*}
%===================================================================

\section{Framework for Calibrating Low-cost Sensors \label{sec_model}}
We propose a novel framework, SenDaL, for efficiently calibrating low-cost sensors for daily life. SenDaL can improve the performance of low-cost sensors in various environments, providing numerous benefits. For instance, by increasing the accuracy of fine-dust sensors, it becomes possible to accurately measure particulate matter in various locations, not limited to air quality monitoring stations. Additionally, employing low-cost and high-accuracy temperature and humidity sensors enables cost-effective monitoring of diverse smart-farm environments. 
In this section, we first describe the data refinement process. Then, we provide a detailed explanation of the overall structure, as well as the training and inference processes of SenDaL.

\subsection{Data Refinement}
Raw data measured from low-cost sensors may suffer from low accuracy and inconsistent measurement intervals due to various causes such as sensor connection error or delay in measurement. However, the refinement process should be as fast as possible in order to be applied to IoT-controlled devices. Therefore, we perform a data refinement process that is simple enough to be suitable for use within our SenDaL framework. This process is illustrated in Figure \ref{refinement}. After collecting raw data from two sensors \(\mathcal{X}\) and \(\mathcal{Y}\), simple moving average (SMA) \cite{ellis2005smarter} and Hodrick-Prescott (HP) \cite{Hodrick_1997} filters are applied to smooth the data and it is synchronized by linear interpolation to be measured at regular time intervals. Then, we generate time-series \(\mathcal{S}\) for \(\mathcal{X}\) and \(\mathcal{Y}\) to train SenDaL.

\subsection{Overall Structure of Framework\label{sec_structure}}
Figure \ref{model} shows the overall structure of our framework, SenDaL. This framework is based on the observation that data collected from sensors tend to remain consistent in most states, with significant changes occurring only in certain situations. Moreover, SenDaL is designed to be as simple as possible for the application of embedded systems with limited processing power, memory, and storage.

The input \(\mathcal{S}\) is embedded into three different layers; a high-accuracy layer \(\mathcal{C}\), a linear (fully-connected) layer \(\mathrm{fc}\), and an embedding layer \(\mathcal{E}\). We refer to the high-accuracy layer as a \textit{component} layer. The embedding layer, \(\mathcal{E}:=\sigma\circ\mathcal{L}\), consists of a linear layer \(\mathcal{L}\) with a sigmoid activation function \(\sigma\). Layer \(\mathcal{E}\) acts as a classifier that distinguishes whether the time-series is stable. Then, the time-series are calibrated by either a linear layer \(\mathrm{fc}:=\mathrm{fc_{out1}}\circ\mathrm{ReLU}\circ\mathrm{fc_{in1}}\) (i.e., fast inference time but less accurate) or a component layer \(\mathcal{C}:=\mathrm{fc_{out2}}\circ\mathcal{C}'\circ\mathrm{fc_{in2}}\) (i.e., slow inference time but more accurate), where ReLU denotes a leakyReLU \cite{Maas_2013} activation function. In SenDaL, the model \(\mathcal{C}'\) can be selected from either an LSTM-based or a Transformer-based model. SenDaL then returns the scalar value \(\hat{x}\), which is a calibrated result of \(x\) through either linear layer \(\mathrm{fc}\) or component layer \(\mathcal{C}\).

\subsection{An Overview of Training and Inference Processes}
In SenDaL, we use different training and inference methods to consider accuracy, inference speed, and energy consumption. The main training and inference strategy is presented in Figure \ref{concept}, expressed as \textit{bottom-up training} and \textit{top-down inference} processes. We first describe the brief concepts of two different processes in this subsection, and then explain the detailed implementation in Sections \ref{sec_training} and \ref{sec_inference}.

First, the \textit{bottom-up training} process, as shown in Figure \ref{concept}(a), is a methodology aimed at achieving levels of accuracy and speed in dynamically switching between linear and component layers. For instance, in scenarios where the time-series data is stable, the computational resources can be saved by avoiding the use of complex layers. On the other hand, if the time-series is unstable, using a more sophisticated layer (i.e., component layer) can lead to improving accuracy. However, the initial embedding layer lacks the ability to make this determination on its own. Therefore, we adopt a training strategy in which calibration results from all layers are propagated to the embedding layer, enabling it to decide which model to use. This process resembles a bottom-up decision-making strategy, wherein the leader integrates the opinions of all members to make the final decision.

Second, the \textit{top-down inference} process, as shown in Figure \ref{concept}(b), is a procedure designed to accomplish fast inference speed and lower energy consumption. In contrast, the bottom-up strategy, which necessitates the utilization of all calibration models, tends to slow down the inference speed. Therefore, the bottom-up strategy is employed solely until the embedding layer is adequately trained to make independent decisions. Once the embedding layer has received sufficient training, it adopts a top-down strategy during the inference process, allowing it to autonomously formulate decisions without relying on any calibration model. Consequently, only the selected calibration model, determined by the embedding layer, is employed for calibrating the time-series, thereby improving latency and energy efficiency. As shown in Figure \ref{concept}(c), the top-down process can be used after the bottom-up process. %\textcolor{red}{\sout{More detailed descriptions of bottom-up training and top-down inference process will be discussed in subsequent Section \ref{sec_training} and Section \ref{sec_inference}, respectively.}}

%===================================================================
\begin{figure*}[th]
\centering
\includegraphics[width=0.8\textwidth]{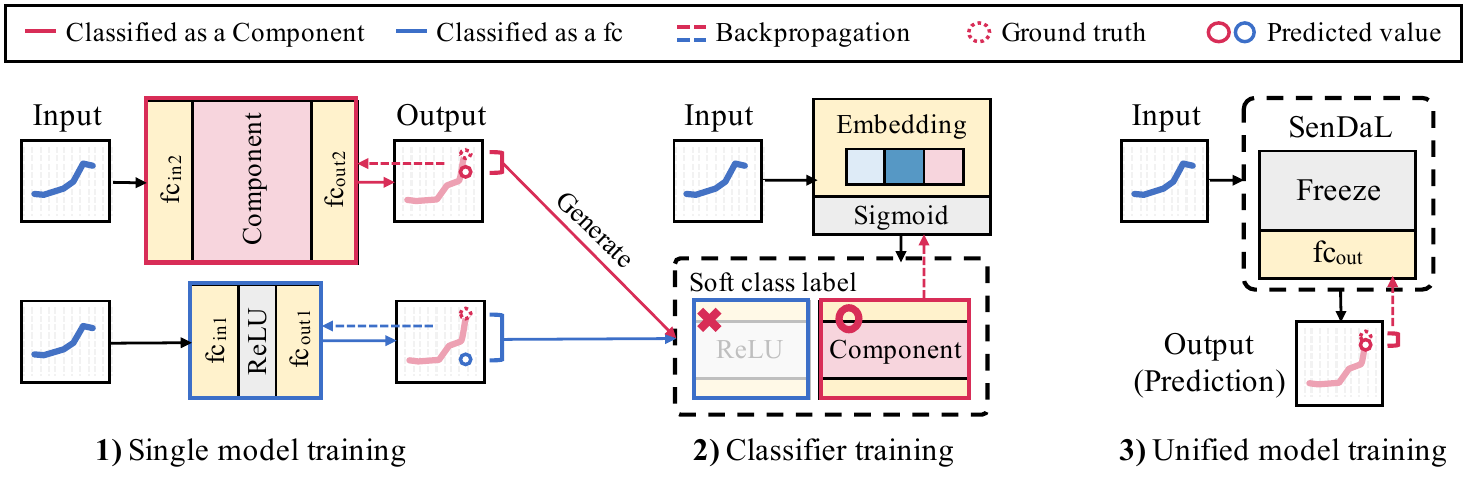} % Reduce the figure size so that it is slightly narrower than the column. Don't use precise values for figure width.This setup will avoid overfull boxes.
\caption{\textbf{The three-step bottom-up training process of SenDaL.} Freeze notation in unified model training denotes a layer in which weight does not change during backpropagation. }
\label{training}
\end{figure*}
%===================================================================

\subsection{Bottom-Up Training\label{sec_training}}

As illustrated in Figure \ref{training}, the bottom-up training process is divided into three main phases; \textbf{1)} \textit{Single model training}: Each calibration layer (i.e., linear and component layer) is trained individually to generate their respective calibration results; \textbf{2)} \textit{Classifier training}: Following the single model training phase, the calibration results from all calibration layers are collected. Then the embedding layer is trained to make independent decisions based on these calibration results; \textbf{3)} \textit{Unified model training}: All layers are fine-tuned to enable top-down inference during the inference process.
%Train each calibration layer and generate their calibration results (Single model training), 2) Collect calibration results from calibration layers and train the embedding layer to make independent decisions (Classifier training), 3) Fine-tune all layers to enable top-down inference during the inference process (Unified model training). Detailed explanations will be provided in each subsequent subsection.
Detailed explanations for each of these phases will be provided in subsequent subsections.

\subsubsection{Single model training\label{sec_singlemodeltraining}}
\noindent Prior to training the embedding layer \(\mathcal{E}\), we first individually train each of the calibration layers, \(\mathrm{fc}\) and \(\mathcal{C}\). Specifically, for \(x_i\) and \(y_i\) corresponding to time-series \(\mathcal{S}_i\), we aim to minimize two losses \(\mathcal{L}_{\mathrm{fc}}\) and \(\mathcal{L}_{\mathcal{C}}\):

\begin{equation}
    \begin{aligned}
        &\mathcal{L}_{\mathrm{fc}}:=\frac{1}{\left|Train\right|}\sum_{i\in Train} \left[y_i-\mathrm{fc}\left(\mathcal{S}_i\right)\right] \\
        &\mathcal{L}_{\mathcal{C}}:=\frac{1}{\left|Train\right|}\sum_{i\in Train} \left[y_i-\mathcal{C}\left(\mathcal{S}_i\right)\right] 
    \end{aligned}    
\end{equation}
where \(Train\) denotes a set of indexes of the time-series used for the training data. These learning results are employed to create labels for layer \(\mathcal{E}\), in the subsequent phase.

\subsubsection{Classifier training}
\noindent In this phase, we train the embedding layer \(\mathcal{E}\). To train embedding layer \(\mathcal{E}\), a soft class label \(y^{*}\) must be generated for the given the time-series \(\mathcal{S}\). These labels are automatically created using the calibration results passed by pre-trained calibration models from the previous phase, which is of the form:
\begin{equation}
    y^{*}_{i}=(\dot{f}\circ\mathcal{N})\left(\bar{y}'_i+w\cdot s_i\right)
\end{equation}
where \(\dot{f}:\left[0, 1\right]\rightarrow\left[0, 1\right]\) denotes a magnifier, \(\mathcal{N}\) denotes a normalizer, \(\bar{y}'_i\) is a smoothed-hard class label, \(w\) is a weighting parameter, and \(s_i\) is a instability factor. Detailed descriptions of some factors are as follows.

\begin{itemize}[noitemsep, leftmargin=1.2em]
\item \noindent\textit{Smoothed-hard class label (\(\bar{y}'_i\))}: The smoothed-hard class label \(\bar{y}'_i\) prevents the hard class labels \(y'_i\) from being too sensitive to make learning difficult. Hard class label \(y'_i\) determines which calibration result of the two layers (i.e., \(\mathrm{fc}\) and  \(\mathcal{C}\)) trained in the previous phase (i.e., single model training) has a smaller error on time-series \(\mathcal{S}_i\):
\begin{equation}\label{Difficulty}
    y'_{i}=
    \begin{cases}
    1, & \mbox{if }\left| y_i-\mathrm{fc}\left(\mathcal{S}_i\right) \right|<\left| y_i-\mathcal{C}\left(\mathcal{S}_i\right) \right| + \xi \\
    0, & \mbox{otherwise}
    \end{cases} \\
\end{equation}
where \(\xi\) denotes the margin of the decision boundary. In this context, \(\xi\) is used to enhance the inference speed of the model by assigning a higher weight to the linear layer. Therefore, \(\xi\geq0\) is considered. Then our smoothed class label \(\bar{y}'_i\) is calculated by:
\begin{equation}\label{Sensibility}
    \bar{y}'_i=
    \begin{cases}
    1, & \textrm{if }\displaystyle\sum\limits_{\substack{i-\lambda <j\leq i \\ j>0}} y'_j <\displaystyle\frac{\lambda}{2} \\
    0, & \mbox{otherwise}
    \end{cases} \\
\end{equation}
where \(\lambda\) represents a smooth factor.

\item \noindent\textit{Instability factor (\(s_i\))}: The instability factor \(s_i\) is defined as a high value if the time-series is unstable, calculating how often the hard class label \(y'_i\) changes, defined by:
\begin{equation}
    s_{i}=\sum_{\substack{i-\lambda <j\leq i \\ j>1}} y'_j \oplus y'_{j - 1}
\end{equation}
where \(\oplus\) is an XOR operator. Weights should be assigned to the component layer as frequent changes of hard class labels can reduce calibration accuracy. The smoothed-hard class label \(\bar{y}'_i\) returns 0 if there are many zeros in the hard class label \(y'_i\) even if it changes frequently, but the instability factor \(s_i\) always returns a higher value only if they change frequently.

\item \noindent\textit{Magnifier (\(\dot{f}\))}: The magnifier \(\dot{f}\) is to magnify small values. In this paper, \(\dot{f}(x)=\frac{(n+1)x}{nx+1}\) with \(n>0\) is used.

\end{itemize}

Then we train the embedding layer \(\mathcal{E}\) to minimize the following binary cross-entropy loss \(\mathcal{L}_\mathrm{class}\):
\begin{equation}
\begin{aligned}
    \mathcal{L}_\mathrm{class}:=\frac{1}{\left|Train\right|}& \sum_{i\in Train}  \left[ y^{*}_i\log\mathcal{E}\left(\mathcal{S}_i\right) \right.\\
     & +\left.\left(1-y^{*}_i\right)\log\left(1-\mathcal{E}\left(\mathcal{S}_i\right)\right)\right]
\end{aligned}
\end{equation}

\subsubsection{Unified model training\label{sec_unifiedmodeltraining}}
\noindent After learning all layers of SenDaL, all of them are combined and trained. The last layers connected to each calibration layer \(\mathrm{fc}\) and \(\mathcal{C}\), are fine-tuned. This enhances the performance of each calibration layer, specialized for situations determined by the embedding layer, enabling top-down inference. Therefore, the two layers \(\mathrm{fc}\) and \(\mathcal{C}\) can achieve more accurate calibration for each case. In other words, we aim to minimize:
\begin{equation}
\begin{aligned}
    \mathcal{L}:=\frac{1}{\left|Train\right|}
    & \biggl[\sum_{\substack{i\in Train \\ \overline{\mathcal{E}}\left(\mathcal{S}_i\right)\leq\theta}} \left(\mathrm{fc_{out1}}\circ\mathrm{ReLU}\circ\overline{\mathrm{fc_{in1}}}\right)\left(\mathcal{S}_i\right)\biggr.\\
    & +  \biggl. \sum_{\substack{i\in Train \\ \overline{\mathcal{E}}\left(\mathcal{S}_i\right)>\theta}} \left(\mathrm{fc_{out2}}\circ\overline{\mathcal{C}'}\circ\overline{\mathrm{fc_{in2}}}\right)\left(\mathcal{S}_i\right)\biggl]
\end{aligned}
\end{equation}
where \(\overline{*}\) (i.e., \(\overline{\mathcal{E}}\), \(\overline{\mathcal{C}'}\), \(\overline{\mathrm{fc_{in1}}}\) and \(\overline{\mathrm{fc_{in2}}}\)) denote the layer \(*\) whose weight does not change, and threshold \(\theta\) is set to 0.5. 

\subsection{Top-Down Inference\label{sec_inference}}
\noindent The training process enables the embedding layer \(\mathcal{E}\) to determine the stability of a given time-series \(\mathcal{S}\), independently of the assistance of the linear and component layers, \(\mathrm{fc}\) and \(\mathcal{C}\). Subsequently, the inference process proceeds by selecting the calibration layer that aligns with the decision made by the embedding layer, in a top-down fashion. As a result, the inference process of our proposed model \(\mathcal{F}\) is as follows:
\begin{equation}\label{United}
    \mathcal{F}(\mathcal{S})=
    \begin{cases}
    \mathcal{C}(\mathcal{S}), & \mbox{if }\mathcal{E}(\mathcal{S}) > \theta \\
    \mathrm{fc}(\mathcal{S}), & \mbox{otherwise}
    \end{cases} \\
\end{equation}
If the condition \(\mathcal{E}(\mathcal{S})\leq \theta\) is satisfied, the input will not be processed by the component layer \(\mathcal{C}\), instead, it will be determined by the linear layer \(\mathrm{fc}\). This is because most time-series are stable, which increases the likelihood of the input being processed by the linear layer \(\mathrm{fc}\). As a result, the inference speed that involves layer \(\mathcal{C}\) is mostly ignored, leading to the conclusion that SenDaL achieves high accuracy with fast inference speed. 

If we assume locality, it is likely that similar variations will appear in the neighboring region. Therefore, it is unnecessary to repeatedly pass through layer $\mathcal{E}$. By adjusting the period through \(\mathcal{E}\), the inference time of \(\mathcal{E}\) is also reduced.

\section{Experiment\label{sec_experiment}}

\subsection{Datasets \label{sec_dataset}}
Since SenDaL represents the first framework for enhancing the performance of low-cost sensors, there are currently no benchmark datasets suitable for assessing its performance. It may be more practical to verify the performance of SenDaL using a diverse range of not fully-refined data collected from real-world sources (e.g., home appliances) rather than relying solely on benchmark datasets. To address this, data were acquired from two types of fine-dust sensors in three different real-world environments for 5-10 days each, labeled Env1, Env2, and Env3. All data were refined at 15 second intervals to a 5 minutes window size. The metadata of our refined datasets is shown in the Table \ref{Metadata}.

We collected a larger number of data compared to most existing studies \cite{Concas_2021} that calibrated data in daily life. We chose a fine-dust sensor for our experiment because it is one of the most popular sensor used in daily life. Additionally, fine-dust data was selected due to its practicality in collecting data over a short period to obtain data for training, as the frequency of variations is higher than that of other sensor data. %While data from other sensors may exhibit different variations compared to the fine-dust sensor, the overall trend of time-series would not be changed significantly. Therefore, if we use the other sensors, not a fine-dust sensor, we should collect a sufficient amount of data over a longer period.}

%===================================================================
\begin{table}[t]

\centering
\caption{\textbf{Details of refined datasets.} The laser type has three attributes; PM10, PM2.5, and PM1, and the infrared type stores the observation of four different sensors. \(m\) denotes expectation, and \(\sigma\) is a standard deviation of the dataset.}
\setlength\tabcolsep{6.5pt}
\resizebox{0.95\columnwidth}{!}{%
\begin{threeparttable}

\begin{tabular}{clccc}
    \toprule
    \textbf{Dataset} & \textbf{Type} & \textbf{Shape} & \textbf{Length} & \textbf{Statistics (\(m\pm\sigma\))} \\
    \midrule
    \multirow{2}{*}{Env1} & Laser & 57270\(\times\)3 & 10\(_{\textrm{days}}\) & 11.892\(\pm\)22.217 \\
        & Infrared & 57270\(\times\)4 & 10\(_{\textrm{days}}\) & \:\:1.743\(\pm\)\:\:2.282 \\
    \midrule
    \multirow{2}{*}{Env2} & Laser & 40883\(\times\)3 & 8\(_{\textrm{days}}\) & \:\:8.952\(\pm\)13.723 \\
        & Infrared & 40883\(\times\)4 & 8\(_{\textrm{days}}\) & \:\:0.991\(\pm\)\:\:2.130 \\
    \midrule
    \multirow{2}{*}{Env3} & Laser & 27697\(\times\)3   & 5\(_{\textrm{days}}\) & 21.433\(\pm\)12.828 \\
        & Infrared & 27697\(\times\)4 & 5\(_{\textrm{days}}\) & \:\:2.539\(\pm\)\:\:2.510 \\
    \bottomrule
\end{tabular}

\begin{tablenotes}

\end{tablenotes}

\end{threeparttable}
}
\label{Metadata}
\end{table}
%===================================================================

\subsection{Experimental Setup \label{sec_setup}}
Various types of fine-dust sensors are available, with beta-ray sensor utilized in air quality monitoring stations being the most accurate. However, the cost of beta-ray sensor makes them unfeasible for use in daily life devices (i.e., home appliances). Instead, we employed two types of fine-dust sensors, specifically infrared (PPD42NS) and laser types (PMS7003), which are commonly utilized in home appliances. There is no specific criterion for the minimum number of these sensors to obtain accurate data, but \cite{Zheng_2018} and \cite{Bulot_2019} claim that when using at least three sensors can yield similar correlation coefficients and slopes. Therefore, we deployed eight PMS7003 laser sensors and four PPD42NS infrared sensors to measure particulates generated in various daily environments, such as turning on/off candles, cooking, laundry, and ventilation. Fine-dust data was collected using Arduino and Raspberry Pi and refined to a 15-second frequency for training purpose.

The laser-type sensors are usually more expensive and accurate than the infrared sensors \cite{Alfano_2020}. Thus, we set the PMS7003 as a high-cost and high-accuracy sensor and the PPD42NS as a low-cost and low-accuracy sensor. The laser-type sensor measures the concentrations of PM1.0, PM2.5, and PM10 separately, whereas the infrared sensor measures fine-dust data without these distinctions. We set the PPD42NS to enable the measurement of find-dust over PM10 in our experiments. Therefore, We used the average PM10 values acquired from the PMS7003 as the ground truth \(y\) to improve the accuracy of the measured value \(x\) from each PPD42NS.

\subsection{Baseline Models\label{sec_baseline}}
Given the limited number of studies using deep learning to calibrate data from low-cost sensors, identifying the widely adopted baseline models can be challenging. Therefore, we employed popular deep learning models as baselines, specifically LSTM-based and Transformer-based models. LSTM-based models were chosen based on a number of citations, and Transformer-based models were selected based on models focusing on resource efficiency. 

As a result, LSTM \cite{Hochreiter_1997}, GRU \cite{Cho_2014}, and Phased LSTM \cite{Neil_2016} were used as LSTM-based models, and Transformer \cite{Vaswani_2017}, Reformer \cite{Kitaev_2020}, and Performer \cite{Choromanski_2021} were utilized as transformer-based models. Additionally, we also adopted Linear regression \cite{freedman_2009} as a baseline, since it is commonly used for calibrating low-cost sensors due to its minimal computational cost. A brief description of the baselines used in this paper is as follows:

\begin{itemize}[noitemsep, leftmargin=1.2em]
\item\noindent\textbf{Linear regression} \cite{freedman_2009}: Linear regression models the relationship between a response and the explanatory variables. 
\item\noindent\textbf{LSTM} \cite{Hochreiter_1997}: LSTM (Long short-term memory) consists of a cell state, a forget gate, an input gate, and an output gate, which is appropriate for time-series prediction.
\item\noindent\textbf{GRU} \cite{Cho_2014}: GRU (Gated recurrent unit) is similar to LSTM but has fewer parameters. GRU often outperforms LSTM on certain smaller datasets.
\item\noindent\textbf{Phased LSTM} \cite{Neil_2016}: Phased LSTM extends the LSTM by adding a time gate, which enables faster convergence. It is suitable for non-synchronized or non-periodic datasets.
\item\noindent\textbf{Transformer} \cite{Vaswani_2017}: The transformer uses a self-attention mechanism and positional encoding, enabling it to weigh the input data and facilitate parallelization.
\item\noindent\textbf{Reformer} \cite{Kitaev_2020}: Reformer uses reversible residual layers and dot-product attention with locality-sensitive hashing, improving performance and efficient memory usage.
\item\noindent\textbf{Performer} \cite{Choromanski_2021}: Performer can estimate full-rank-attention Transformers while using only linear space and time complexity, without the need to make any assumptions.
\end{itemize}

\subsection{Evaluation Methods\label{sec_evaluation}}
We employed an anchored walk-forward optimization \cite{nousi2019machine} to evaluate the performance of the model using 10-fold cross-validation. The performance of the model was determined by averaging the results obtained from each low-cost sensor. To verify the superiority of SenDaL, various evaluation methods were utilized. A description of the evaluation metrics used in this paper is as follows:

\begin{itemize}[noitemsep, leftmargin=1.2em]
\item\noindent\textbf{RMSE (Root mean squared error)}: 
RMSE is a measure of the difference of predicted value \(\hat{x}\) and the ground truth \(y\) using mean squared error (i.e., 
\(\sqrt{\mathbb{E}_n{[{\left(\hat{x}-y\right)^{\scalebox{1}{$\scriptscriptstyle 2$}}}]}}\) where \(\mathbb{E}_n\) is an (empirical) expectation with \(n\) test data). This metric is widely used for fine-dust sensor calibration to evaluate overall accuracy \cite{Alfano_2020}.
\item\noindent\textbf{Miss ratio}: We also employ an auxiliary accuracy metric for specific cases: the miss ratio \cite{Michael_2021}. This metric denotes the proportion of instances where the error \(\varepsilon\) between the predicted value \(\hat{x}\) and the ground truth \(y\) exceeds a threshold \(\theta_\textbf{M}\) (i.e., \(\varepsilon:=\left|\hat{x}-y\right|\geq\theta_\textbf{M}\)) to ensure the reliability and consistency of correction results. RMSE is inappropriate to measure the accuracy in cases where correction results temporarily exhibit significant errors. Hence, we utilize the miss ratio as an auxiliary accuracy metric.
\item\noindent\textbf{Latency}: 
Latency is the average CPU inference speed, which can be empirically calculated as the average time difference between \(t_{\mathrm{start}}\) and \(t_{\mathrm{end}}\) as it passes through the model (i.e., \(\mathbb{E}_n[t_{\mathrm{end}}-t_{\mathrm{start}}]\)).
\item\noindent\textbf{Average energy consumption (E\(_{\textrm{CPU}}\), E\(_{\textrm{RAM}}\))}: We calculate average CPU energy consumption (E\(_{\textrm{CPU}}\)) and average RAM energy consumption (E\(_{\textrm{RAM}}\)) to consider the practical computational speeds in embedded environments. Since these metrics are quite ambiguous to empirically estimate, we measured both energy consumption based on the CodeCarbon\footnote{https://pypi.org/project/codecarbon/} library. Here, the RAM power usage is calculated based on 3 Watts per 8GB memory, and the CPU power usage is derived from the value tracked in the intel RAPL\footnote{https://www.intel.com/content/www/us/en/developer/topic-technology/software-security-guidance/advisory-guidance.html} (Running Average Power Limit) interface.
\end{itemize}

Especially, the last three evaluation metrics (i.e., Latency, E\(_{\textrm{CPU}}\), and E\(_{\textrm{RAM}}\)) assess the feasibility of the model when applied to embedded hardware using low-cost sensors. For each evaluation metric, the value was determined by calculating the average of five measurements for reliability. Also, we focused on evaluating these metrics in CPU-only environments since embedded hardware used in home appliances typically only uses the CPU.
Additionally, we evaluated the model inference speed by converting to ONNX\footnote{https://onnx.ai/} (Open Neural Network Exchange) and FlatBuffers\footnote{https://flatbuffers.dev/} serialization format for the reliability of the experiment.

\subsection{Implementation Details\label{sec_implementation}}
We implemented and evaluated all models using PyTorch, with some models making use of external libraries, such as Performer\footnote{https://pypi.org/project/performer-pytorch/} and Reformer\footnote{https://pypi.org/project/reformer-pytorch/}. 
The window size \(N\) was fixed at 20 to ensure a 5-minute input window length. The hidden size of LSTM-based and Transformer-based models utilized in the experiment was set to 16 and 8, respectively. The models were trained on a machine with an AMD Ryzen 5 5600X 6-Core processor and an NVIDIA GeForce RTX 2060 graphics card.

%===================================================================
\begin{table*}[t]
\centering
\caption{\textbf{Main experimental results on three different datasets} in terms of RMSE, average CPU inference latency (Latency), energy consumption on RAM (E\(_{\textrm{RAM}}\)) and on CPU (E\(_{\textrm{CPU}}\)), respectively. * denotes a model implemented by external libraries, hence it may not be fully optimized compared to the model implemented by PyTorch API in terms of latency and energy consumption.}
\resizebox{1\textwidth}{!}{%
\begin{threeparttable}
\setlength\tabcolsep{6pt}
\begin{tabular}{llccccccccccccccccccc}
    \toprule

    %\midrule

    \multirow{2}{*}{\textbf{Component}} & \multirow{2}{*}{\textbf{Model}} & 
    \multicolumn{4}{c}{\textbf{RMSE\scriptsize$\downarrow$}} & \multicolumn{4}{c}{\textbf{Latency\scriptsize$\downarrow$} (\(\mu\)s)} & \multicolumn{4}{c}{\textbf{E\(_{\textrm{RAM}}\)\scriptsize$\downarrow$} (hJ)} & \multicolumn{4}{c}{\textbf{E\(_{\textrm{CPU}}\)\scriptsize$\downarrow$} (hJ)} \\ 
    \cmidrule(lr){3-6} \cmidrule(lr){7-10} \cmidrule(lr){11-14} \cmidrule(lr){15-18}
       %&& RMSE & Latency\tnote{a} & E\(_{\textrm{RAM}}\)\tnote{b} & E\(_{\textrm{CPU}}\)\tnote{c}
       && Env1 & Env2 & Env3 & \textbf{Avg}
       & Env1 & Env2 & Env3 & \textbf{Avg}
       & Env1 & Env2 & Env3 & \textbf{Avg}
       & Env1 & Env2 & Env3 & \textbf{Avg}\\
    
    \midrule

    Linear regression & Vanilla & 2.297 & 2.366 & 3.110 & 2.617 & 0.023 & 0.023 & 0.023 & 0.023 & 0.124 & 0.126 & 0.123 & 0.124 & 0.236 & 0.243 & 0.247 & 0.242
    \\

    \midrule
    \multirow{2}{*}{LSTM} & Vanilla & 2.233 & 2.320 & 3.054 & 2.562 & 0.398 & 0.397 & 0.388 & 0.394 & 2.342 & 2.360 & 2.370 & 2.357 & 4.246 & 4.281 & 4.305 & 4.278
    
    \\
    & \textbf{SenDaL} & \textbf{2.195} & \textbf{2.298} & \textbf{3.038} & \textbf{2.538} & \textbf{0.061} & \textbf{0.054} & \textbf{0.060} & \textbf{0.058} & \textbf{0.393} & \textbf{0.411} & \textbf{0.399} & \textbf{0.401} & \textbf{0.719} & \textbf{0.755} & \textbf{0.738} & \textbf{0.737} \\

    \midrule
    \multirow{2}{*}{GRU} & Vanilla & 2.299 & \textbf{2.064} & 3.053 & 2.508 & 0.404 & 0.401 & 0.444 & 0.416 & 2.387 & 2.420 & 2.456 & 2.421 & 4.328 & 4.390 & 4.462 & 4.394 \\
    & \textbf{SenDaL} & \textbf{2.243} & 2.065 & \textbf{3.004} & \textbf{2.471} &  \textbf{0.064} & \textbf{0.057} & \textbf{0.081} & \textbf{0.067} & \textbf{0.401} & \textbf{0.476} & \textbf{0.422} & \textbf{0.433} & \textbf{0.733} & \textbf{0.873} & \textbf{0.780} & \textbf{0.795} \\

    \midrule
    \multirow{2}{*}{Phased LSTM*} & Vanilla & \textbf{2.282} & 2.072 & \textbf{2.995} & \textbf{2.450} & 1.432 & 1.439 & 1.444 & 1.414 & 8.238 & 8.780 & 11.421 & 9.479 & 14.926 & 15.915 & 20.695 & 17.178 \\
    & \textbf{SenDaL} & 2.286 & \textbf{2.033} & 3.031 & \textbf{2.450} & \textbf{0.082} &  \textbf{0.084} & \textbf{0.089} & \textbf{0.085} & \textbf{0.402} & \textbf{0.470} & \textbf{0.581} & \textbf{0.484} & \textbf{0.734} & \textbf{0.865} & \textbf{1.079} & \textbf{0.892} \\

    \midrule
    \multirow{2}{*}{Transformer} & Vanilla & 3.632 & 3.064 & \textbf{2.920} & 3.220 & 0.540 & 0.532 & 0.584 & 0.552 & 3.284 & 3.177 & 3.191 & 3.217 & 5.955 & 5.767 & 5.802 & 5.841 \\
    & \textbf{SenDaL} & \textbf{2.199} & \textbf{2.324} & 3.059 & \textbf{2.556} & \textbf{0.076} & \textbf{0.069} & \textbf{0.081} & \textbf{0.075} & \textbf{0.503} & \textbf{0.483} & \textbf{0.524} & \textbf{0.503} & \textbf{0.920} & \textbf{0.889} & \textbf{0.973} & \textbf{0.927} \\

    \midrule
    \multirow{2}{*}{Reformer*} & Vanilla & 2.387 & 3.314 & \textbf{3.122} & 2.968 & 1.874 & 1.768 & 1.935 & 1.859 & 10.188 & 9.707 & 9.976 & 9.957 & 18.457 & 17.592 & 18.072 & 18.040 \\
    & \textbf{SenDaL} & \textbf{2.304} & \textbf{2.281} & 3.150 & \textbf{2.610} & \textbf{0.104} & \textbf{0.057} & \textbf{0.876} & \textbf{0.346} & \textbf{0.401} & \textbf{0.426} & \textbf{2.579} & \textbf{1.135} & \textbf{0.732} & \textbf{0.782} & \textbf{4.684} & \textbf{2.066} \\
    
    \midrule
    \multirow{2}{*}{Performer*} & Vanilla & 2.514 & 2.481 & \textbf{3.041} & 2.691 & 0.815 & 0.759 & 0.868 & 0.771 & 4.848 & 4.868 & 4.321 & 4.679 & 8.782 & 8.821 & 7.839 & 8.481 \\
    & \textbf{SenDaL} & \textbf{2.260} & \textbf{2.202} & 3.119 & \textbf{2.562} & \textbf{0.080} & \textbf{0.069} & \textbf{0.319} & \textbf{0.169} & \textbf{0.386} & \textbf{0.398} & \textbf{1.035} & \textbf{0.606} & \textbf{0.706} & \textbf{0.730} & \textbf{1.890} & \textbf{1.109} \\

    \bottomrule
\end{tabular}

% \begin{tablenotes}
% \footnotesize{\item[a] Average inference latency (\(\mu\)s)}
% \footnotesize{\item[b] Energy consumption on RAM (0.1\(\mu\)J)}
% \footnotesize{\item[c] Energy consumption on CPU (0.1\(\mu\)J)}
% \end{tablenotes}
\end{threeparttable}
}
\label{result_main}
\end{table*}
%===================================================================

\subsection{Experimental Results\label{sec_result}}
Table \ref{result_main} shows the overall performance of SenDaL in our experiment. SenDaL shows outstanding performance on all evaluation metrics, including RMSE, average CPU latency, and energy consumption on RAM and CPU. Detailed discussions are introduced in the following subsections.

\subsubsection{Accuracy} As shown in Table \ref{result_main}, SenDaL achieves higher accuracy than each individual component in most models and environments. This implies that SenDaL does not rely solely on the linear layer and the component layer individually, but rather learns to extract the unique features of time-series data from low-cost sensors considering their trends. One might infer that the performance of SenDaL would be an average of the two models since it simply combines linear and component models. However, as the results have demonstrated, SenDaL approaches or surpasses the accuracy of the individual components. This implies that SenDaL does not simply select between the linear and component layers, but rather has a similar effect to an ``ensemble" through the bottom-up training process. Consequently, SenDaL selectively uses the component model only in cases of unstable time-series, which can lead to an increase in accuracy. We then evaluated the performance of each component. In contrast to the results obtained in previous studies related to time-series \cite{Vaswani_2017}, the accuracy of the Transformer-based models was lower than that of the LSTM-based models. This can be attributed to the high level of noise present in the data collected by low-cost sensors. The LSTM-based models utilize the correction result of the previous data as a prediction, while the Transformer-based models employ an attention technique. Removing all the noise is challenging since low-cost sensors have low measurement accuracy even after various preprocessing steps. As a result, the Transformer-based models are more susceptible to incorporating noise into the model. Thus, it can be inferred that the LSTM-based models effectively capture the trend of the data from low-cost sensors more accurately than the Transformer-based models.

%===================================================================
\begin{table}[t]
\centering
\caption{\textbf{Calibration consistency compared to the linear model.} Linear models often have faults that differ significantly from the ground truth, making them less reliable even if the RMSE is similar. SenDaL is comparable in speed to linear models, yet its miss ratio is similar to deep learning models. }
\resizebox{0.95\columnwidth}{!}{%
\begin{threeparttable}
\setlength\tabcolsep{7.4pt}
\begin{tabular}{lccccccc}
    \toprule

    \multirow{2}{*}{\textbf{Model}} & \multirow{2}{*}{\(\mathbf{\theta}_\textbf{M}\)} & 
    \multicolumn{5}{c}{\textbf{Miss ratio\scriptsize$\downarrow$}} \\ 
    \cmidrule(lr){3-7} && Env1 & Env2 & Env3 & \textbf{Avg} & \textbf{Impv.}\tnote{*}\\
    
    \midrule
    Linear regression & \multirow{3}{*}{3} & 0.487 & 0.438 & \textbf{0.674} & 0.512 & -
    \\
    Vanilla LSTM && 0.456 & 0.489 & 0.683 & 0.517 & -
    \\
    \textbf{SenDaL-LSTM} && \textbf{0.454} & \textbf{0.435} & 0.677 & 
    
    \textbf{0.497} & \tcbhighmath[blue2 eqbox]{\textcolor{BlueViolet}{\scriptsize{\textbf{+3.0\%}}}}
    \\

    \midrule
    
    Linear regression & \multirow{3}{*}{5} &  0.293 & 0.321 & 0.543 & 0.357 & -
    \\
    Vanilla LSTM && \textbf{0.274} & 0.308 & \textbf{0.528} & \textbf{0.341} & -
    \\
    \textbf{SenDaL-LSTM} && \textbf{0.274} & \textbf{0.302} & 0.546 & 0.343 & 
    \tcbhighmath[blue3 eqbox]{\textcolor{BlueViolet}{\scriptsize{\textbf{+4.1\%}}}}
    \\

    \midrule
    Linear regression & \multirow{3}{*}{10} & 0.060 & 0.186 & 0.292 & 0.152 & -
    \\
    Vanilla LSTM && 0.066 & \textbf{0.146} & 0.299 & 0.143 & -
    \\
    \textbf{SenDaL-LSTM} && \textbf{0.053} & 0.164 & \textbf{0.291} & \textbf{0.142} & 
    \tcbhighmath[blue4 eqbox]{\textcolor{BlueViolet}{\scriptsize{\textbf{+7.0\%}}}}
    \\
    \midrule
    Linear regression & \multirow{3}{*}{30} & 0.026 & \textbf{0.000} & 0.095 & 0.033 & -
    \\
    Vanilla LSTM && 0.023 & 0.008 & \textbf{0.041} & \textbf{0.022} & -
    \\
    \textbf{SenDaL-LSTM} && \textbf{0.022} & \textbf{0.000} & 0.057 & 0.023 & 
    \tcbhighmath[blue5 eqbox]{\textcolor{BlueViolet}{\scriptsize{\textbf{+43.5\%}}}}
    \\

    \bottomrule
\end{tabular}

\begin{tablenotes}
\small{\item[*] Rate of decrease in miss ratio compared to linear models.}
\end{tablenotes}
\end{threeparttable}
}
\label{result_reliability}
\end{table}
%===================================================================

\subsubsection{Efficiency} As shown in Table \ref{result_main}, SenDaL demonstrated significant improvement over vanilla models in terms of latency and energy consumption when evaluated using only CPU. A particularly significant difference was observed when the component model was complex. For example, the latency of SenDaL with GRU was 7 times faster than that of the vanilla model and 20 times faster for Phased LSTM. Similarly, energy consumption also demonstrated a consistent trend. This difference depends not only on the model but also on the dataset. For instance, Performer is 10 times faster than Env1 and Env2, but only 3 times faster than Env3. This is because the proportion of the deep learning model usage varies across each model and environment. The worst-case scenario is when deep learning models are employed in all situations, which can result in slower performance than vanilla models. However, assigning appropriate weight to linear models can greatly improve efficiency in terms of speed and energy consumption, as it does not require the use of deep learning models in all cases. Consequently, SenDaL is an efficient model that approximates the latency and energy consumption of linear models without compromising model inference accuracy.

\subsubsection{Consistency and reliability}
Some individuals may infer from Table \ref{result_main} that a linear model is more practical than deep learning models due to their faster inference speeds and lower energy consumption. It can be considered that while the RMSE of the linear model is only slightly lower when compared to deep learning models, the average difference in calibration error between the two models is only 0.1, which is considered to be sufficiently negligible. However, it should be noted that RMSE is an average error, and thus it does not account for some large variations in data, particularly in cases where the data is mostly stable. To address this issue, a comparison of the miss ratio between the linear and deep learning models. As demonstrated in Table \ref{result_reliability}, the miss ratio of the linear model is consistently higher than that of the deep learning model, particularly as the error threshold increases. For instance, when the threshold is set to 30, the miss ratio of the linear model is 44\% higher compared to that of the deep learning model. This suggests that linear models may not be as reliable when dealing with large changes, as the errors incurred are significantly greater than those of deep learning models. In contrast, SenDaL has been shown to improve the miss ratio, making it comparable to or even lower than that of the deep learning model. This is a significant finding, as it indicates that SenDaL provides consistent calibration results even under conditions of significant variability. Thus, SenDaL can be considered a reliable approach for low-cost sensor calibration, as it not only guarantees improved reliability when dealing with large changes but also offers similar speeds and energy consumption as compared to the linear model.

%===================================================================
\begin{table*}[t]
\centering
\caption{\textbf{Average CPU inference latency (\(\mu\)s) of PyTorch and ONNX model on different embedded hardware.} Phased LSTM and Reformer were excluded from the ONNX experiment since they were not implemented to be optimized for ONNX operators.}
\resizebox{\textwidth}{!}{%
\begin{threeparttable}
\setlength\tabcolsep{4.25pt}

\begin{tabular}{cllccccccccccccccc}
    \toprule

    % \multicolumn{17}{c}{AA}%Unit=Float32} 
    % \midrule

    \multirow{2}{*}{\textbf{Format}} & \multirow{2}{*}{\textbf{Component}} & \multirow{2}{*}{\textbf{Model}}  & \multicolumn{5}{c}{\textbf{Raspberry Pi 3\tnote{a}}} & \multicolumn{5}{c}{\textbf{Raspberry Pi 4\tnote{b}}} & 
    \multicolumn{5}{c}{\textbf{Jetson Nano\tnote{c}}} \\ 
    \cmidrule(lr){4-8} \cmidrule(lr){9-13} \cmidrule(lr){14-18} 
       &&& Env1 & Env2 & Env3 & \textbf{Avg} & \textbf{Impv.}
       & Env1 & Env2 & Env3 & \textbf{Avg} & \textbf{Impv.}   & Env1 & Env2 & Env3 & \textbf{Avg}  & \textbf{Impv.}  \\
    
    \midrule

    \multirow{15}{*}{PyTorch} & Linear regression & Vanilla & 0.226 & 0.221 & 0.232 & 0.227 & - & 0.194 & 0.191 & 0.194 & 0.193 & - & 0.275 & 0.284 & 0.275 & 0.278 & - \\

    \cmidrule(lr){2-18}
    &\multirow{2}{*}{LSTM} & Vanilla & 3.781 & 3.665 & 3.859 & 3.768 & - & 3.209 & 3.201 & 3.242 & 3.218 & - & 3.635 & 3.786 & 3.718 & 3.713 & - \\
    && \textbf{SenDaL} 
    & \textbf{0.659} & \textbf{0.559} & \textbf{0.650} & \textbf{0.623} & 
    \tcbhighmath[blue4 eqbox]{\textcolor{BlueViolet}{\scriptsize{\textbf{+605\%}}}} 
    & \textbf{0.467} & \textbf{0.416} & \textbf{0.488} & \textbf{0.457} &
    \tcbhighmath[blue4 eqbox]{\textcolor{BlueViolet}{\scriptsize{\textbf{+704\%}}}} & \textbf{0.659} & \textbf{0.633} & \textbf{0.654} & \textbf{0.649} & 
    \tcbhighmath[blue4 eqbox]{\textcolor{BlueViolet}{\scriptsize{\textbf{+572\%}}}} 
    \\

    \cmidrule(lr){2-18}
    &\multirow{2}{*}{GRU} & Vanilla & 3.582 & 3.565 & 3.568 & 3.572 & - & 3.091 & 3.039 & 3.112 & 3.081 & - & 3.600 & 3.359 & 3.507 & 3.488 & - \\
    & & \textbf{SenDaL}
    & \textbf{0.641} & \textbf{0.569} & \textbf{0.702} & \textbf{0.637} & 
    \tcbhighmath[blue4 eqbox]{\textcolor{BlueViolet}{\scriptsize{\textbf{+560\%}}}} 
    & \textbf{0.473} & \textbf{0.419} & \textbf{0.550} & \textbf{0.481} & 
    \tcbhighmath[blue4 eqbox]{\textcolor{BlueViolet}{\scriptsize{\textbf{+641\%}}}} & \textbf{0.673} & \textbf{0.599} & \textbf{0.721} & \textbf{0.664} & 
    \tcbhighmath[blue4 eqbox]{\textcolor{BlueViolet}{\scriptsize{\textbf{+525\%}}}} 
    \\

    \cmidrule(lr){2-18}
    &\multirow{2}{*}{Phased LSTM} & Vanilla & 14.467 & 14.430 & 14.435 & 14.451 & - & 9.552 & 9.474 & 9.766 & 9.597 & - & 10.447 & 10.324 & 10.432 & 10.401 & -  \\
    && \textbf{SenDaL} 
    & \textbf{0.778} & \textbf{0.564} & \textbf{0.750} & \textbf{0.698} & 
    \tcbhighmath[blue5 eqbox]{\textcolor{BlueViolet}{\scriptsize{\textbf{+2070\%}}}} 
    & \textbf{0.603} & \textbf{0.416} & \textbf{0.588} & \textbf{0.526} & 
    \tcbhighmath[blue5 eqbox]{\textcolor{BlueViolet}{\scriptsize{\textbf{+1824\%}}}} & \textbf{0.750} & \textbf{0.596} & \textbf{0.602} & \textbf{0.649} & 
    \tcbhighmath[blue5 eqbox]{\textcolor{BlueViolet}{\scriptsize{\textbf{+1603\%}}}} 
    \\

    \cmidrule(lr){2-18}
    &\multirow{2}{*}{Transformer} & Vanilla & 9.718 & 9.697 & 9.647 & 9.687 & - & 20.311 & 18.491 & 19.528 & 19.443 & - & 17.098 & 17.419 & 16.817 & 17.098 & - \\
    && \textbf{SenDaL}
    & \textbf{0.754} & \textbf{0.639} & \textbf{0.955} & \textbf{0.783} & 
    \tcbhighmath[blue5 eqbox]{\textcolor{BlueViolet}{\scriptsize{\textbf{+1287\%}}}}
    & \textbf{0.706} & \textbf{0.418} & \textbf{1.069} & \textbf{0.731} & 
    \tcbhighmath[blue5 eqbox]{\textcolor{BlueViolet}{\scriptsize{\textbf{+2660\%}}}} & \textbf{0.771} & \textbf{0.604}  & \textbf{1.070}  & \textbf{0.815} & 
    \tcbhighmath[blue5 eqbox]{\textcolor{BlueViolet}{\scriptsize{\textbf{+2097\%}}}}
    \\

    \cmidrule(lr){2-18}
    &\multirow{2}{*}{Reformer} & Vanilla & 14.741 & 15.373 & 15.301 & 15.138 & - & 7.805 & 8.163 & 7.923 & 7.963 & - & 10.757 & 10.621 & 10.596 & 10.658 & -  \\
    && \textbf{SenDaL} 
    & \textbf{0.972} & \textbf{0.807} & \textbf{7.457} & \textbf{3.079} & 
    \tcbhighmath[blue3 eqbox]{\textcolor{BlueViolet}{\scriptsize{\textbf{+492\%}}}} 
    & \textbf{0.475} & \textbf{0.371} & \textbf{3.398} & \textbf{1.415} & 
    \tcbhighmath[blue3 eqbox]{\textcolor{BlueViolet}{\scriptsize{\textbf{+563\%}}}} & \textbf{0.897} & \textbf{0.729} & \textbf{4.265} & \textbf{1.967} & 
    \tcbhighmath[blue3 eqbox]{\textcolor{BlueViolet}{\scriptsize{\textbf{+542\%}}}} 
    \\
    
    \cmidrule(lr){2-18}
    &\multirow{2}{*}{Performer} & Vanilla & 7.287 & 6.633 & 7.434 & 4.118 & - & 3.475 & 3.647 & 3.706 & 3.609 & - & 5.115 & 5.086 & 5.088 & 5.096 & - \\
    && \textbf{SenDaL} 
    & \textbf{0.890} & \textbf{0.760} & \textbf{3.646} & \textbf{1.765} & 
    \tcbhighmath[blue3 eqbox]{\textcolor{BlueViolet}{\scriptsize{\textbf{+403\%}}}}
    & \textbf{0.415} & \textbf{0.377} & \textbf{1.812} & \textbf{0.868} & 
    \tcbhighmath[blue3 eqbox]{\textcolor{BlueViolet}{\scriptsize{\textbf{+415\%}}}} & \textbf{0.839} & \textbf{0.753} & \textbf{2.233} & \textbf{1.275} & 
    \tcbhighmath[blue3 eqbox]{\textcolor{BlueViolet}{\scriptsize{\textbf{+400\%}}}}
    \\

    \midrule

    \multirow{11}{*}{ONNX} & Linear regression & Vanilla & 0.283 & 0.282 & 0.283 & 0.283 & - & 0.060 & 0.060 & 0.060 & 0.060 & - & 0.082 & 0.083 & 0.083 & 0.083 & -  \\

    \cmidrule(lr){2-18}
    &\multirow{2}{*}{LSTM} & Vanilla & 0.577 & 0.579 & 0.580 & 0.579 & - & 0.107 & 0.107 & 0.107 & 0.107 & - & 0.140 & 0.140 & 0.140 & 0.140 & -  \\
    && \textbf{SenDaL} & \textbf{0.384} & \textbf{0.376} & \textbf{0.382} & \textbf{0.381} & \tcbhighmath[blue2 eqbox]{\textcolor{BlueViolet}{\scriptsize{\textbf{+52\%}}}} & \textbf{0.079} & \textbf{0.078} & \textbf{0.079} & \textbf{0.079} & \tcbhighmath[blue2 eqbox]{\textcolor{BlueViolet}{\scriptsize{\textbf{+35\%}}}} & \textbf{0.108} & \textbf{0.107} & \textbf{0.108} & \textbf{0.108} & \tcbhighmath[blue2 eqbox]{\textcolor{BlueViolet}{\scriptsize{\textbf{+30\%}}}}  \\

    \cmidrule(lr){2-18}
    &\multirow{2}{*}{GRU} & Vanilla & 0.573 & 0.573 & 0.574 & 0.574 & - & 0.107 & 0.107 & 0.107 & 0.107 & - & 0.145 & 0.147 & 0.146 & 0.146 & -  \\
    && \textbf{SenDaL} & \textbf{0.384} & \textbf{0.382} & \textbf{0.393} & \textbf{0.386} & \tcbhighmath[blue2 eqbox]{\textcolor{BlueViolet}{\scriptsize{\textbf{+48\%}}}} & \textbf{0.079} & \textbf{0.078} & \textbf{0.081} & \textbf{0.079} & \tcbhighmath[blue2 eqbox]{\textcolor{BlueViolet}{\scriptsize{\textbf{+35\%}}}} & \textbf{0.108} & \textbf{0.108} & \textbf{0.110} & \textbf{0.109} & \tcbhighmath[blue2 eqbox]{\textcolor{BlueViolet}{\scriptsize{\textbf{+34\%}}}}  \\

    \cmidrule(lr){2-18}
    &\multirow{2}{*}{Transformer} & Vanilla & 2.279 & 2.332 & 2.317 & 2.310 & - & 0.459 & 0.475 & 0.509 & 0.481 & - & 0.540 & 0.447 & 0.447 & 0.448 & -  \\
    && \textbf{SenDaL} & \textbf{0.412} & \textbf{0.384} & \textbf{0.454} & \textbf{0.417} & \tcbhighmath[blue3 eqbox]{\textcolor{BlueViolet}{\scriptsize{\textbf{+554\%}}}} & \textbf{0.086} & \textbf{0.079} & \textbf{0.096} & \textbf{0.087} & \tcbhighmath[blue3 eqbox]{\textcolor{BlueViolet}{\scriptsize{\textbf{+552\%}}
    }}& \textbf{0.113} & \textbf{0.108}  & \textbf{0.120}  & \textbf{0.114} & \tcbhighmath[blue3 eqbox]{\textcolor{BlueViolet}{\scriptsize{\textbf{+393\%}}}}  \\

    \cmidrule(lr){2-18}
    &\multirow{2}{*}{Performer} & Vanilla & 2.615 & 2.626 & 2.632 & 2.625 & - & 0.448 & 0.471 & 0.488 & 0.469 & - & 0.560 & 0.559 & 0.553 & 0.557 & - \\
    && \textbf{SenDaL} & \textbf{0.410} & \textbf{0.375} & \textbf{1.379}& \textbf{0.722} & \tcbhighmath[blue3 eqbox]{\textcolor{BlueViolet}{\scriptsize{\textbf{+364\%}}}} & \textbf{0.085} & \textbf{0.079} & \textbf{0.264} & \textbf{0.143} & \tcbhighmath[blue3 eqbox]{\textcolor{BlueViolet}{\scriptsize{\textbf{+329\%}}}} & \textbf{0.113} & \textbf{0.107} & \textbf{0.305} & \textbf{0.175} & \tcbhighmath[blue3 eqbox]{\textcolor{BlueViolet}{\scriptsize{\textbf{+318\%}}}}  \\

    \bottomrule
\end{tabular}

\begin{tablenotes}
\small{\item[a] On 1.4GHz 4-core ARM Cortex-A53}
\small{\item[b] On 1.5GHz 4-core ARM Cortex-A72}
\small{\item[c] On 1.43GHz 4-core ARM Cortex-A57 (CPU Only)}
\end{tablenotes}
\end{threeparttable}
}

\label{result_hardware}
\end{table*}
%===================================================================

\subsubsection{Performance on embedded hardware} The results of our experiments demonstrate the feasibility of implementing SenDaL on various embedded hardware platforms, including the Raspberry Pi 3, Raspberry Pi 4, and Jetson Nano, as presented in Table \ref{result_hardware}. Two model formats, PyTorch and ONNX, were employed to conduct the experiments and facilitate a reliable analysis of performance on embedded hardware. 

%\textcolor{red}{Furthermore, we verify the superiority of SenDaL using Arduino Nono, a microcontroller with resources similar to those of actual home appliances, to demonstrate that SenDaL can be employed in real-life home appliances, as shown in Table \ref{result_mcu}. To adapt SenDaL to Arduino Nano, we convert the models to FlatBuffers to ensure compatibility with the C++ platform and assess their performance on the microcontroller.}

%===================================================================
\begin{table}[t]
\centering
\caption{\textbf{Average CPU inference latency (ms) on a microcontroller.} }
\resizebox{0.95\columnwidth}{!}{%
\begin{threeparttable}
\setlength\tabcolsep{6.5pt}

\begin{tabular}{llccccc}
    \toprule

    \multirow{2}{*}{\textbf{Component}} & \multirow{2}{*}{\textbf{Model}} & 
    \multicolumn{5}{c}{\textbf{Arduino Nano 33 BLE Sense\tnote{a}}} \\ 
    \cmidrule(lr){3-7} 
       && Env1 & Env2 & Env3 & \textbf{Avg} & \textbf{Impv.}
       \\
    
    \midrule

    Linear regression & Vanilla & 0.010 & 0.010 & 0.011 & 0.010 & - \\

    \midrule
    \multirow{2}{*}{LSTM} & Vanilla & 2.962 & 2.919 & 2.971 & 2.951 & - \\
    & \textbf{SenDaL} & \textbf{0.080} & \textbf{0.031} & \textbf{0.159} & \textbf{0.090} & \tcbhighmath[blue5 eqbox]{\textcolor{BlueViolet}{\scriptsize{\textbf{+3279\%}}}} \\

    \midrule
    \multirow{2}{*}{Transformer} & Vanilla & 9.047 & 9.051 & 9.054 & 9.051 & -  \\
    & \textbf{SenDaL} & \textbf{0.177} & \textbf{0.031}  & \textbf{0.417}  & \textbf{0.208} & \tcbhighmath[blue5 eqbox]{\textcolor{BlueViolet}{\scriptsize{\textbf{+4351\%}}}} \\

    \bottomrule
\end{tabular}

\begin{tablenotes}
\small{\item[a] On 64MHz ARM Cortex M4 with 1MB Flash memory}
\end{tablenotes}
\end{threeparttable}
}
\label{result_mcu}
\end{table}
%===================================================================

First, the results obtained from different PyTorch models show that SenDaL exhibits significantly lower average CPU inference latency compared to the vanilla models and approaches latency levels close to linear models across all hardware platforms. Moreover, the latency gap between SenDaL and the vanilla model becomes even more prominent when using more complex components. These experiment's findings suggest that implementing SenDaL on embedded hardware, coupled with low-cost sensors commonly used in home appliances, would enable its practical application in daily life. 

Second, similar results were obtained when conducting the same experiment after converting to ONNX format. Most models focus on high-performance GPU environments and often overlook various low-cost environments. For instance, the conversion process for Phased LSTM and Reformer presents considerable challenges. Phased LSTM results in a noticeable decline in performance and Reformer encounters unsupported operators in ONNX format. However, SenDaL demonstrates remarkable efficiency even after conversion to the ONNX format. Specifically, using the ONNX format allows for the application of deep learning on low-cost hardware where machine learning libraries like PyTorch may not be available.

Lastly, we observed consistent results when representative deep learning models (i.e., LSTM and Transformer) were converted to FlatBuffers for compatibility with a microcontroller, specifically Arduino Nano, which has resources similar to those of actual home appliances, as shown in Table \ref{result_mcu}. In the microcontroller environment, deploying the deep learning models posed challenges due to limited memory capacity, and the models operated at a significantly slower speed compared to other hardware environments. Nevertheless, SenDaL demonstrated an inference speed much closer to that of the linear model.

These findings indicate that SenDaL is a practical solution for embedded systems and strongly supports its potential use in real-world applications.

\subsubsection{Flexibility} As shown in Tables \ref{result_main}-\ref{result_mcu}, SenDaL outperforms the vanilla model in most results in terms of accuracy, latency, and energy consumption. Additionally, SenDaL has the capability to create various structures by combining different components, such as LSTM, Transformer, and other models. Therefore, SenDaL can be applied in a wide range of IoT environments, depending on the model's compatibility with hardware resources. For example, Phased LSTM can be used as a component because it has the highest accuracy when hardware resources are sufficient. However, It may be challenging to apply it to low-resource hardware owing to insufficient memory. On the other hand, using GRU ensures fast latency as it has fewer parameters, while its latency is comparable to that of a linear model. Therefore, GRU is more suitable than other models in low-resource hardware. In conclusion, SenDaL can provide high performance that is suitable for various IoT environments' requirements.

\section{Discussion\label{sec_discussion}}

SenDaL exhibits fast inference speed and high calibration accuracy across diverse hardware requirements and scenarios by flexibly adjusting its components. However, it is important to acknowledge certain limitations in our study. First, additional experiments in various environments and scenarios are necessary to continuously validate SenDaL's superiority. For example, more diverse scenarios can be considered by collecting different types of sensor data, such as temperature and humidity sensors. However, the collection of data from these sensors is time-consuming due to the slow rate of change in the data. Despite the time investment, these studies have the potential to demonstrate the reliability of our framework across a broader spectrum of applications. Future work should extend over longer periods to thoroughly develop these datasets.
% Collecting these datasets concurrently with other future works is advisable for enhanced efficiency and effectiveness.

Second, the reliability of the ground truth should be addressed, given that the high-cost sensor used in our experiment may not be sufficiently expensive. Using outdoor data sources, such as weather stations or established benchmarks, could be most accurate as ground truth. However, it is challenging to apply them in our study since our primary focus is calibrating scenarios from daily life. Therefore, the collection of data directly through the use of more expensive sensors may be a focus for future work.

Thirdly, SenDaL employs a methodology involving training the model on the local server and embedding the pre-trained weights in the model. This approach is considered essential when applying and operating SenDaL with real home appliances. To tackle this challenge, options include implementing periodic communication between IoT products and servers or adjusting weights through firmware updates. Therefore, we suggest specifying this process to ensure regular retraining or maintenance of the model.

Finally, the framework faces a challenge in low-cost embedded environments due to its dependence on the entire set of training parameters for the model. This reliance constrains the utilization of larger-scale deep learning models, for instance, those surpassing the available memory constraints to achieve higher accuracy. Consequently, it is imperative to devise a specialized model architecture tailored for daily devices. We suggest considering the design of a model structure specifically crafted for daily applications as a focus for future research.

\section{Conclusion\label{sec_conclusion}}

In this study, we propose SenDaL, a calibration framework for low-cost sensors embedded in IoT devices that offers high accuracy, fast inference speed, and cost-effectiveness. SenDaL utilizes a bottom-up training process with refined time-series data to train the ground truth. As a result, SenDaL outperforms baseline deep learning models in terms of both accuracy and efficiency. Furthermore, SenDaL demonstrates superior performance on embedded hardware, showing its potential for use in IoT home appliances.

For future work, we aim to extend the application of SenDaL to different types of low-cost sensors, including temperature and humidity sensors. Additionally, we will prepare a variety of high-performance measuring instruments to further enhance the accuracy of the ground truth. Finally, we will design a structure optimized for real-world daily devices and apply SenDaL to actual home appliances.

% {\appendix[Proof of the Zonklar Equations]
% Use $\backslash${\tt{appendix}} if you have a single appendix:
% Do not use $\backslash${\tt{section}} anymore after $\backslash${\tt{appendix}}, only $\backslash${\tt{section*}}.
% If you have multiple appendixes use $\backslash${\tt{appendices}} then use $\backslash${\tt{section}} to start each appendix.
% You must declare a $\backslash${\tt{section}} before using any $\backslash${\tt{subsection}} or using $\backslash${\tt{label}} ($\backslash${\tt{appendices}} by itself
%  starts a section numbered zero.)}

\balance


\bibliographystyle{IEEEtran}
\bibliography{IEEEabrv, main}

\newpage

 

% \end{IEEEbiography}



%\vfill

\end{document}